# ACT-GAN: Radio map construction based on generative adversarial networks with ACT blocks[1]

Qi Chen[a], Jingjing Yang[a, b, *], Ming Huang [a, b,*], and Qiang Zhou [a]
[a]*School of Information, Yunnan University, Kunming, PR China*
[b]*Key Laboratory of Spectral Sensing and Radio Monitoring, Yunnan University, Kunming, PR China*

**Abstract.** The radio map, serving as a visual representation of electromagnetic spatial characteristics, plays a pivotal role in assessment of wireless communication networks and radio monitoring coverage. Addressing the issue of low accuracy existing in the current radio map construction, this paper presents a novel radio map construction method based on generative adversarial network (GAN) in which the Aggregated Contextual-Transformation (AOT) block, Convolutional Block Attention Module (CBAM), and Transposed Convolution (T-Conv) block are applied to the generator, and we name it as ACT-GAN. It significantly improves the reconstruction accuracy and local texture of the radio maps. The performance of ACT-GAN across three different scenarios is demonstrated. Experiment results reveal that in the scenario without sparse discrete observations, the proposed method reduces the root mean square error (RMSE) by 14.6% in comparison to the state-of-the-art models. In the scenario with sparse discrete observations, the RMSE is diminished by 13.2%. Furthermore, the predictive results of the proposed model show a more lucid representation of electromagnetic spatial field distribution. To verify the universality of this model in radio map construction tasks, the scenario of unknown radio emission source is investigated. The results indicate that the proposed model is robust radio map construction and accurate in predicting the location of the emission source.

Keywords: Radio map construction, generative adversarial network, sparse sampling, robustness

## 1. Introduction

Radio map characterizes the field distribution information of electric waves in electromagnetic space, which can be expressed in terms of electric field distribution, power spectral density (PSD) distribution, and path loss (PL). It is a quantitative description of electromagnetic wave propagation from multiple dimensions such as time, spectrum, space, and power. In the field of wireless communications, the construction of radio maps is closely related to wireless communication network planning and radio management [5-8; 17; 19]. Additionally, radio map finds applications in the fields of wireless localization [3], unmanned aerial vehicle (UAV) path planning [4; 18; 20; 28], and autonomous driving [10; 14]. Therefore, the pursuit of constructing a high-precision radio map construction is currently a research hotspot.

The common radio map construction approach is to use the PSD values of the known locations to interpolate the PSD values of the unknown location. These methods include Inverse Distance Weighted (IDW) interpolation [12], Kriging interpolation [1], Radial Basis Function (RBF) interpolation [11], and Model-Based Interpolation (MBI) [27]. Since these algorithms are primarily based on sampling, they often neglect the underlying principles of radio wave propagation. Consequently, they do not require the prior knowledge of physical parameters, such as the location, number and power of emission sources, making them suitable for rapid estimation of maps in open electromagnetic space environments. However, the accuracy of these methods is directly proportional to the quantity of samples collected, and the large amount of sample collection leads to significant time and cost expenditures. Another common approach is to calculate path loss by directly modeling the area to be predicted, including empirical models [30], dominant path models [22], and more complex ray tracing models [16]. The main limitations of these methods lie in their sensitivity to unknown noise, high computational complexity when predicting at numerous locations, and a lack of generalization capability.

[1] This research was funded by the National Natural Science Foundation of China (62361055, 61963037, 62261059) and the Fifteenth Graduate Student Research and Innovation Program of Yunnan University (KC-23234401)
*Corresponding author. Jingjing Yang, E-mail: yangjingjing@ynu.edu.cn.; Ming Huang, E-mail: Huang Ming huangming@ynu.edu.cn.

In recent years, the increasing development in the field of artificial intelligence has provided robust support for radio map construction research. Unlike traditional methods, these approaches leverage the potent data-fitting capabilities of artificial neural networks to autonomously map from input information to output PSD measurements values. Deep learning models are currently used in radio map construction. For instance, the data generated by the propagation model was utilized by Teganya [21] et al. to train a fully convolutional auto-encoder, and a RMSE of 2dB is obtained. A skip-connection convolutional model named RadioUnet was proposed by Levie R [13] et al., and its ability to directly reconstruct PSD distribution maps from urban building distribution maps was demonstrated. Additional MBI information was incorporated into the RadioUnet model by Zhang [27] et al. to train a generative adversarial network and a better prediction results was achieved. The other deep learning methods used for radio map construction include reinforcement learning [26], transfer learning [15], and Gaussian processes [23]. These methods have the advantages of short time-consuming high robustness and generalization ability, which are the mainstream prediction methods. However, there are some prominent challenges with the existing models: the relatively low prediction accuracy, the unclear image textures, and the limited application scenario.

To address the above problems, this paper designs a deep learning model (ACT-GAN) in which the aggregated contextual-transformation (AOT) [25], Convolutional Block Attention Module (CBAM) [24], and the Transposed Convolution (T-Conv) block are applied to the generator for radio map construction. Based on the *RadioMapSeer* public dataset [13], radio maps were constructed for three diverse application scenarios. The experimental results show that the radio map reconstruction accuracy of the proposed model in three different scenes is improved by an average of 11.8% compared to the current optimal model. Interestingly, in the scenario of unknown radio emission source the proposed model improves the emission source localization accuracy by 34.3% compared with the existing model. In addition, the reconstruction quality is significantly improved, indicating that the proposed method is versatile in multi-scene and multi-tasking. We open-sourced our codes of the ACT-GAN model at *https://github.com/YNUniversityCQ/ACT-GAN.* to promote research on radio map construction and emission source localization.

## 2. Radio map construction process and scenario assumptions

### 2.1. Construction process

The common process of radio map construction is illustrated in Figure 1, which includes three stages: data collection, algorithm analysis, and radio map construction. In this paper, we will conduct experiments on three different scenarios and demonstrate the applicability of the proposed radio map construction method.

### 2.2. Scenario assumptions

Scenario 1: It is assumed that both the transmitter's location and the distribution of buildings are known, and the goal is to train the neural network for predicting propagation path loss from the transmitter position vector (x) to any receiving point position vector (y). This scenario is mainly used for wireless communication network planning and coverage prediction.

Scenario 2: In anticipation of the demands for autonomous driving and UAV path planning, future radio map construction will be carried out in complex spatiotemporal environments, where discrete and sparse observations are obtained by deploying sensing devices. In order to simulate this kind of scenario, sparse discrete observations are introduced as additional input features.

Scenario 3: The location of the emission source is unknown, and the radio map is constructed with sparse PSD sampling points and the location of the emission source is estimated. This scenario is applicable to the field of radio monitoring, where illegal emission source needs to be located with limited receivers.

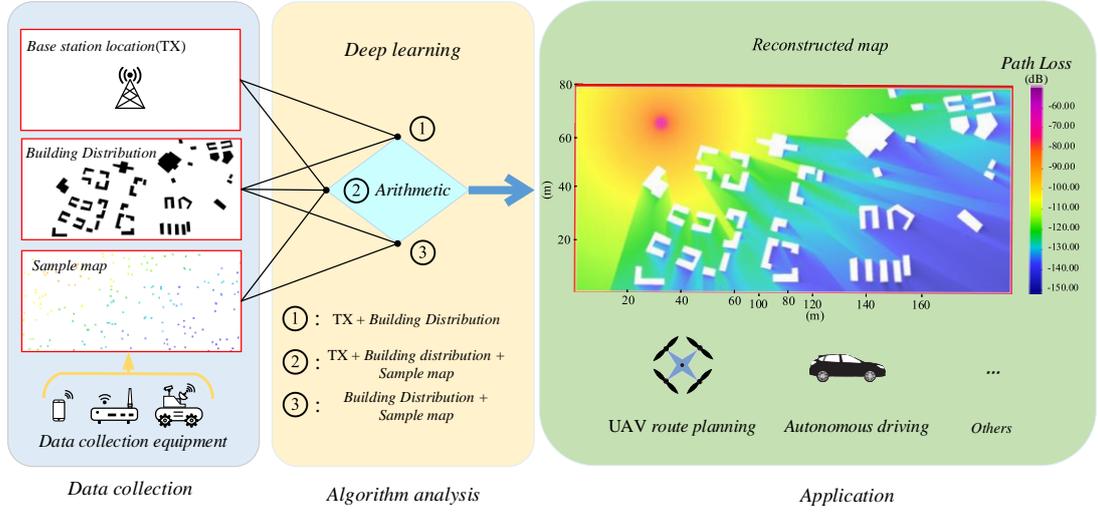

Fig.1. Radio map construction process.

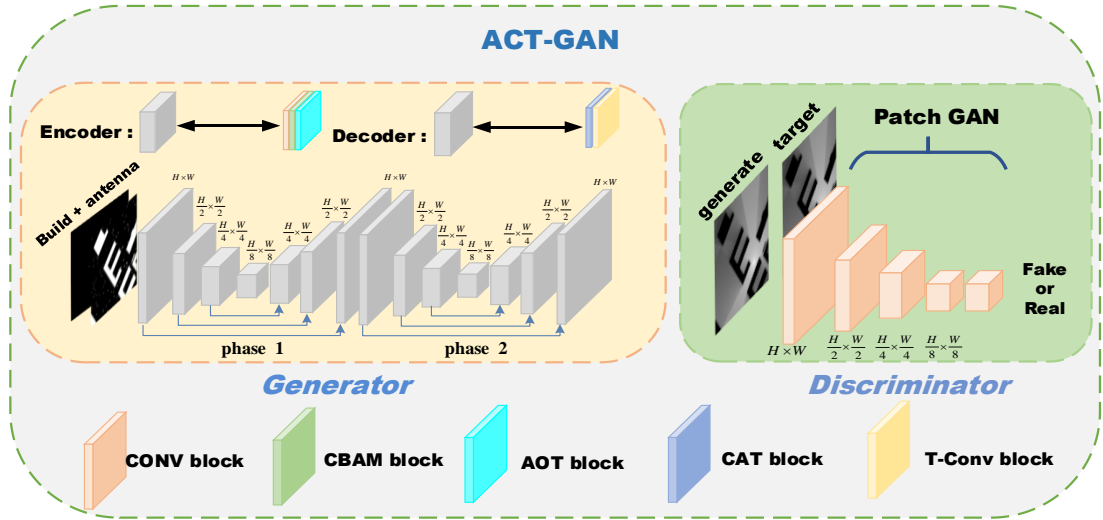

Fig.2. The structure of ACT-GAN model.

## 3. Model

### 3.1. ACT-GAN structure

The structure of the proposed ACT-GAN model is depicted in Figure 2, which consists of an enhanced Generator and a Discriminator.

For the Discriminator model, the fully convolutional Patch-GAN, as proven in [9] and [29], has shown superior image translation performance. Hence, we adopt the Patch-GAN as the Discriminator model.

Generator models are generally designed based on the encoder-decoder structure. A well-designed encoder-decoder structure is crucial in determining the performance of a convolutional neural network. Therefore, this paper introduces an innovative generator structure and employs a two-stage simultaneous training strategy to update the gradient parameters.

### 3.2. Encoder

The function of the encoder is to encode the features of the input data and map it to a high-dimensional space. In our proposed model, each encoder layer consists of a concatenated series of a convolution (CONV) block, an Attention Mechanism Block (Convolutional Block Attention Module, CBAM), and an AOT block, as illustrated in Fig.3.

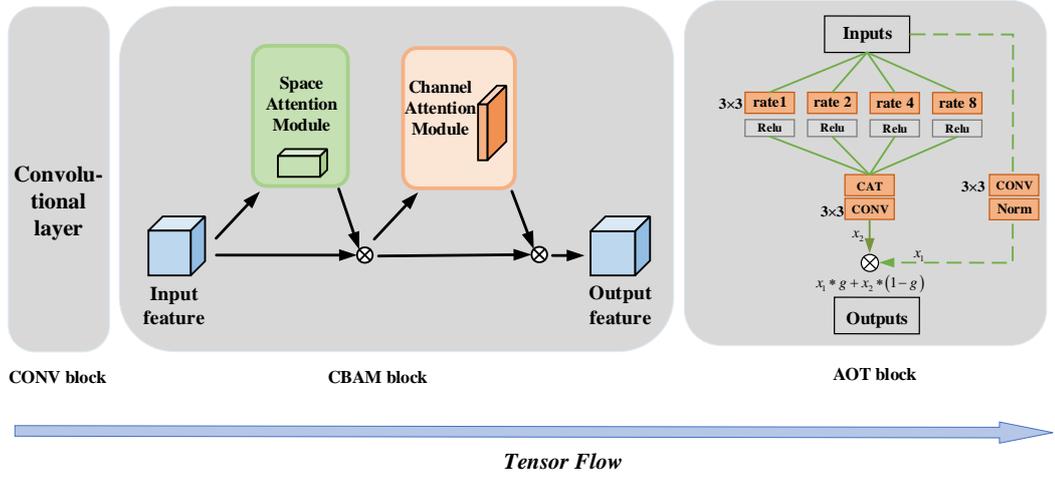

Fig.3. The structure of the Encoder.

In this structure, the CONV block is used for feature encoding and downsampling, the CBAM block is used to promote the network to learn the channel of interest or the location of the area of interest, and the AOT block captures multi-scale contextual information from the image. Within the encoder structure, the addition of the AOT block is the key to ensure that the constructed radio map has a clear texture. In general, the new designed encoder can better extract map features and perceive long-range information interactions.

As shown in Figure 3, the AOT block adopts the split-transformation- aggregate strategy from top to bottom. 1)Splitting: the AOT block splits the standard convolution kernel into four sub-kernels, with each output channel being one-quarter of the original input channels. 2)Transforming: the four sub-kernels are transformed using dilated convolutions with dilation rates of 1, 2, 4, and 8 on the split channels. Sub-kernels with larger dilation rates acquire a broader receptive field to enhance global accuracy in radio map construction. Sub-kernels with smaller dilation rates, on the other hand, focus more on local electromagnetic spatial neighborhood information to enhance local textures. 3)Aggregating: the contextual information from different receptive fields is finally aggregated through concatenation and subsequently undergoes feature fusion via standard convolution. Overall, this module can be used to better control the global information and local texture features during radio map construction. The final output of the module is shown in Eq. (1), where $g$ is an adaptive hyperparameter.

$$out = x_2 \times g + x_1(1-g) \qquad (1)$$

### 3.3. Decoder

The function of the decoder is to map features from high-dimensional space to low-dimensional space, enabling feature decoding. In the proposed model, each decoder layer consists of a concatenate block (CAT block) and a transposed convolution block (T-Conv block), all designed to recover semantic information of each layer. Here, the CAT block stacks channel features between the encoder and the decoder. The T-Conv block is used to recover features into image and perform upsampling, which is less likely to produce blurred images compared to bidirectional interpolation methods.

Generator of the ACT-GAN model contains two stages, and each has 7 encoding-decoding layers, where the parameters of the CONV block are shown in Table 1. Additionally, the AOT block and CBAM block are attached after the convolution layers, sharing the same resolution and number of channels.

Table 1
Generator convolution layer structure parameters

| Layer | input | 1 | 2 | 3 | 4 | 5 | 6 | 7 | output |
|---|---|---|---|---|---|---|---|---|---|
| Resolution | 256 | 256 | 128 | 64 | 32 | 64 | 128 | 256 | 256 |
| Channel | 2or3 | 64 | 128 | 256 | 512 | 256 | 128 | 64 | 1 |
| kernel | - | 7 | 4 | 4 | 4 | 3 | 3 | 3 | - |

## 3.4. Loss function

In order to capture the global information and local texture of the radio map, four loss functions including construction loss, style loss, perceptual loss, and adversarial loss of Patch-GAN are used in optimal control.

Firstly, the optimization objective is to minimize the Mean Square Error (MSE) to ensure a high level of global accuracy for the radio map, and the loss function is depicted as

$$L_{MSE} = \frac{1}{N} \cdot \sum_{i=1}^{N} \|y_i - \hat{y}_i\|^2, \qquad (2)$$

where $y_i$ and $\hat{y}_i$ is the true and estimated pixel values, respectively, and $N$ is the total number of pixels.

Secondly, since the effectiveness of perceptual loss and style loss has been widely proven in the field of computer vision, this paper incorporates them to enhance local texture. The purpose of perceptual loss is to minimize the $L_1$ distance between the predicted image and label, while the style loss is defined as the $L_1$ distance between the Gram matrix of deep features in the predicted map and label, thus, the loss functions are expressed as Eqs. (3) and (4), respectively.

$$L_{per} = \sum_i \frac{\|\phi_i(y) - \phi_i(\hat{y})\|_1}{N_i} \qquad (3)$$

$$L_{sty} = \mathbb{E}_i \left[ \left\| \phi_i \; y^T \phi_i \; y - \phi_i \; \hat{y}^T \phi_i \; \hat{y} \right\|_1 \right]. \qquad (4)$$

Here, $\phi_i$ is the activation map from the $i$ th layers of a pretrained network in VGG19 [2], $N_i$ is the number of elements in $\phi_i$, $\mathbb{E}$ is the mathematical expectation.

Finally, the adversarial loss of Patch-GAN is added to improve the visual fidelity of the constructed radio map, and the overall optimization function can be defined as

$$L = \lambda_{MSE} L_{MSE} + \lambda_{per} L_{per} + \lambda_{sty} L_{sty} + \lambda_{adv} L_{adv}^G. \qquad (5)$$

## 4. Experimental verification

### 4.1. Dataset

The performance of the ACT-GAN model is validated on the RadioMapSeer dataset, which consists of 700 maps, each with 80 transmitter locations, for a total of 56,000 simulated radio maps. The city maps are taken from OpenStreetMap, and the heights of the transmitters, receivers, and buildings are set to 1.5m, 1.5m, and 25m, respectively.

During the training process, the loss function hyperparameters are set to be $\lambda_{adv} = 0.01$, $\lambda_{MSE} = 1$, $\lambda_{per} = 0.1$ and $\lambda_{sty} = 250$, respectively. All training and testing processes are conducted under the Pytorch deep learning framework on an NVIDIA GeForce RTX 3090 GPU. The initial learning rate is set to $1 \times 10^{-4}$. After 50 epochs of training, the learning rate is reduced by a factor of 10 to fine-tune the model parameters, resulting in a total of 100 epochs. During training, a VGG19 model pre-trained on ImageNet is used to compute the style loss and perceptual loss.

### 4.2. Evaluation metrics

To evaluate the performance of ACT-GAN, we use two error metrics: the RMSE and the normalized mean square error (NMSE), defined as

$$RMSE = \frac{1}{N} \sum_{i=1}^{N} \sqrt{\frac{\sum_{j=1}^{M} \left( \hat{r}^{(i,j)} - r^{(i,j)} \right)^2}{M}}, \qquad (6)$$

$$NMSE = \frac{1}{N} \sum_{i=1}^{N} \frac{\sum_{j=1}^{M} \left( \hat{r}^{(i,j)} - r^{(i,j)} \right)^2}{\sigma(r^{(i)})}. \qquad (7)$$

Here, N and M represent the width and height of the radio map resolution, $r^{(i,j)}$ denotes the true receiving field strength (RSS) value for grid $(i,j)$, and $\hat{r}^{(i,j)}$ represents the predicted RSS value for grid $(i,j)$. $\sigma(r^{(i)})$ represents the variance of the true RSS value in the i-th grid.

## 4.3. Result

Performance of the proposed ACT-GAN is compared with the other methods in three scenarios. The whole dataset comprises 700 regions. For each scenario, the dataset is split to train data, validation data and test data with 500 regions, 100 regions and 100 regions, respectively.

### 4.3.1. Scenario 1

The path loss threshold is the critical coefficient for judging the background noise of the radio map. $(\delta)_{dB} = 10\log_{10}(W \cdot N_0) + NF$ represents the background noise in decibels (dB), where $W$, $N_0$, and $NF$ are bandwidth, thermal noise PSD, and noise coefficient respectively. As shown in Eq. (8), in the region where $PL_{ture} < PL_{thr}$, the path loss values are truncated and the pixel values are set to zero; otherwise, the pixel values are transformed by $\frac{PL_{ture} - PL_{thr}}{1 - thr}$, where $PL_{ture}$ is the true path loss

$$PL_{thr} = -(P_{Tx})_{dB} + SNR_{thr} + (\delta)_{dB}, \quad (8)$$

where $(P_{Tx})_{dB}$ is the path loss value at the emission source, and $SNR_{thr}$ is the signal-to-noise ratio (SNR) below the threshold. Radio maps under four different thresholds are shown in Figure 4.

It can be seen from Figure 4 that as the threshold increases, the image color shows a trend of becoming lighter in gray value. This is due to the continuous increase of the background noise, which causes large-scale fading in the radio wave propagation process. In contrast, for deep learning models, it will be easier to achieve convergence because the neural network will be more sensitive to regions with larger grayscale values.

Considering the real background noise intensity in the urban environment, three small thresholds are set for experiments in this scenario. Only two features, building distribution and transmitter location information, are used as input. The performances of RadioUNet, Deep Autoencoder (Deep AE), and Radial Basis Function Neural Network (RBF) are compared under three different thresholds. The RMSE of the test set is shown in Figure 5.

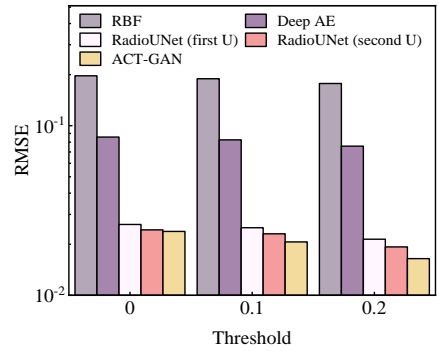

Fig.5. RMSE under different thresholds. (The first U and second U corresponds to the first and second training stage of RadioUNet, respectively.)

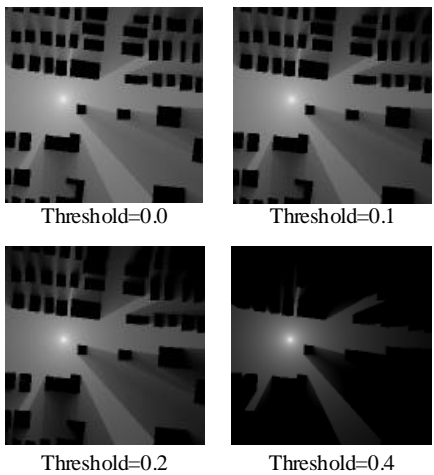

Fig.4. Radio maps under different thresholds.

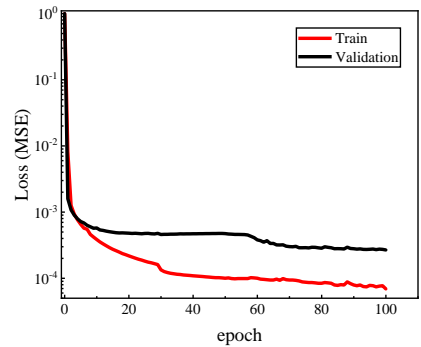

Fig.6. Loss curves at different stages.

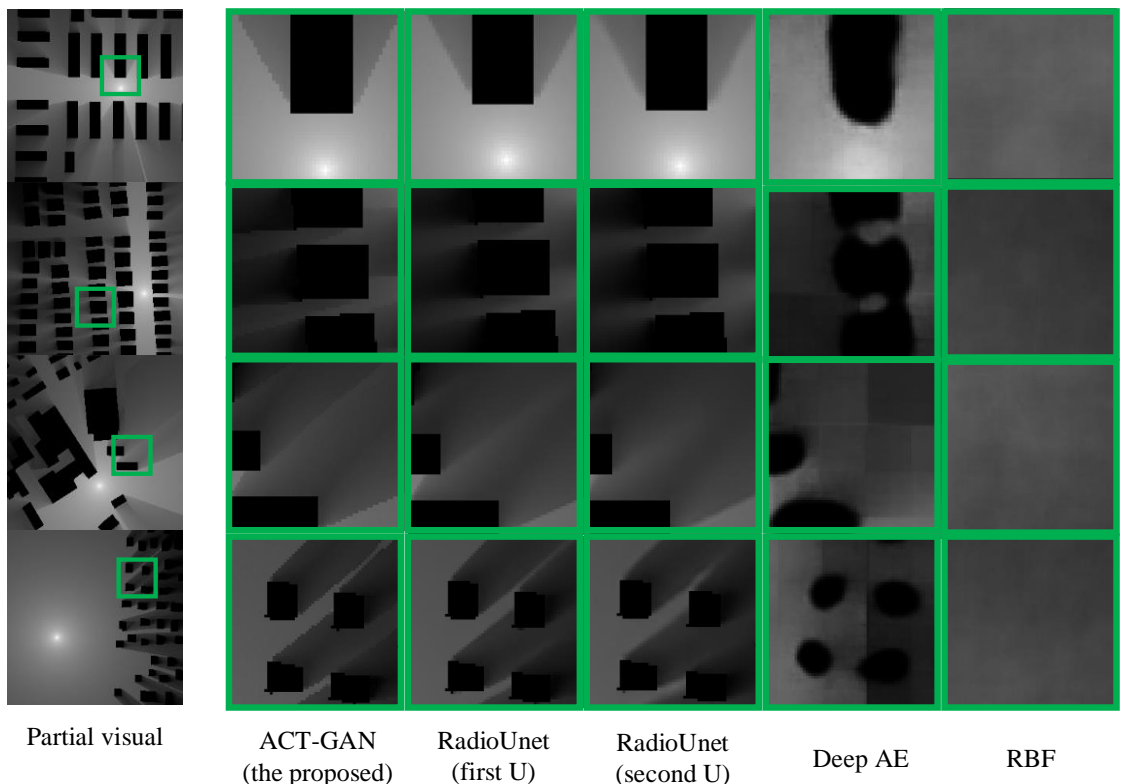

| Partial visual | ACT-GAN (the proposed) | RadioUnet (first U) | RadioUnet (second U) | Deep AE | RBF |

Fig.7. Local textures for different models at different stages.

From Figure 5, it can be observed that the proposed model performs better at smaller thresholds. Moreover, at the general threshold of 0.2, it achieves a 14.6% reduction in RMSE compared to the optimal model (RadioUNet) reported in [13]. In order to prevent overfitting during training, we also calculate the verification loss, and the trend of the loss over time for both the training and validation process is shown in Figure 6. The MSE of both the training set and the validation set tend to stabilize after approximately 30 epochs.

Due to the multipath effect and the impact of the reflection, diffraction and scattering from buildings on wave propagation, it increases the difficulty of predicting path loss. The RMSE only reflects the global accuracy of the constructed map unilaterally. As shown in Figure 7, four randomly selected radio maps from the test set are shown on the left, where the green rectangular region is locally zoomed to the right. It can be seen that the ACT-GAN model shows a clearer texture in terms of the quality of the map components, i.e., it is more consistent with the real radio propagation phenomenon.

Table 2

Comparison of RMSE and NMSE under settings *a*, *b,* and *c*

| Model | Setting a | | Setting b | | Setting c | |
| --- | --- | --- | --- | --- | --- | --- |
| | NMSE | RMSE | NMSE | RMSE | NMSE | RMSE |
| Kriging | 0.8836* | 0.1947* | 0.8987* | 0.1962* | 1.0146 | 0.1968 |
| RBF | 0.3532* | 0.1343* | 0.1830* | 0.0884* | 0.6687 | 0.1670 |
| Deep AE | 0.1898* | 0.0998* | 0.3152* | 0.1295* | 0.0336 | 0.0349 |
| Unet | 0.0093* | 0.0220* | 0.0053* | 0.0166* | 0.0038 | 0.0179 |
| RadioUnet | 0.0052* | 0.0164* | 0.0042* | 0.0148* | 0.0035 | 0.0171 |
| RME-GAN | **0.0043*** | 0.0151* | 0.0036* | 0.0130* | - | - |
| ACT-GAN | 0.0046 | **0.0133** | **0.0032** | **0.0111** | **0.0035** | **0.0118** |

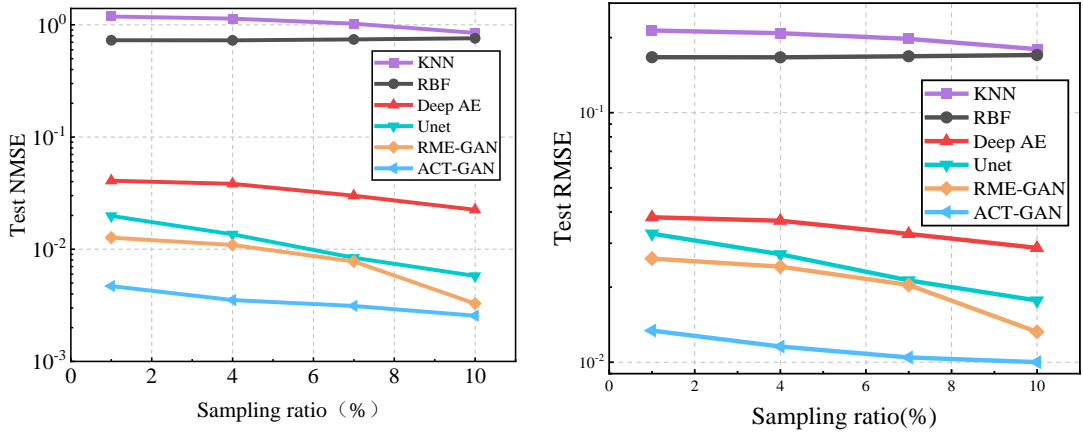

Fig.8. Comparison of NMSE (left) and RMSE (right) at different sampling rates.

### 4.3.2. Scenario 2

On the basis of scenario 1, this scenario adds sparse PSD measurement values under three different settings: *a*, *b*, and *c*, as another one-dimensional input for radio map construction.

Setting *a* (Uniformly distributed): 1% of sparse observations from the whole radio map are uniformly sampled.

Setting *b* (Unbalanced sample distribution among regions): The radio map with a ratio ranging from 1% to 10% in each region is randomly sampled, and the sparse observations are distributed uniformly.

Setting *c* (Non-uniformly distributed): Similar to Setting *b*, but the sparse observation values are randomly distributed.

Additionally, we also conduct uniform sampling at fixed sampling rates of 4%, 7%, and 10% respectively, and compared five different models. The results are shown in Figure 8.

As shown in Figure 8, the ACT-GAN model outperforms the other models in terms of RMSE and NMSE indicators at various sampling rates. In addition, the estimation accuracies under three different experimental settings are shown in Table 2.

From Table 2 (data marked with * is quoted from [27]), it is evident that the ACT-GAN achieves higher accuracy in constructing radio maps under all three different settings. Furthermore, it is worth noting that setting *b* and *c* with uniformly distributed sparse observations are more likely to obtain higher construction accuracy. But in practice, collecting uniformly distributed sparse observations is challenging, while randomly distributed sparse observations are more universal in various scenarios.

### 4.3.3. Scenario 3

In the scenario where the location and number of the emission sources are unknown, the purpose of radio map construction is to estimate the location of emission sources. It may have potential application in radio monitoring.

When conducting sparse sampling, achieving radio map construction with fewer sampling points and fewer sampling frequencies are two key issues. To address the first issue, we conduct experiments with different numbers of sampling points, labeled as $\Omega=20$, $\Omega=50$, $\Omega=100$, and $\Omega=150$. For the second issue, after selecting the sampling locations, we extend 3 meters (3 pixels) in both the horizontal and vertical directions, and finally get a $4\times 4$ square meter sampling area as a single sampling block. This sampling method reduces the logarithmic path loss correlation between each sampling point and can better reflect the performance of the prediction model. A comparison of RMSE and NMSE for different sampling points is given in Table 3.

From Table 3, it can be observed that the ACT-GAN model outperforms the other models under four different numbers of sampling points. Considering the complexity of the radio wave propagation, robustness experiments are conducted under the sampling points of $\Omega=100$. Specifically, Gaussian random noise with different standard deviations ($\sigma=10$, $\sigma=20$, $\sigma=50$) are added to simulate the real radio propagation process. Results under the influence of Gaussian noise with different standard deviations are shown in Figure 9. It is seen that the ACT-GAN model shows stronger robustness under different noise intensities, and it can still ensure a high construction quality even in a high-noise environments ($\sigma=50$).

Table 3
Comparison of different models at different sampling points.

| Model | $\Omega=20$ | | $\Omega=50$ | | $\Omega=100$ | | $\Omega=150$ | |
|---|---|---|---|---|---|---|---|---|
| | RMSE | NMSE | RMSE | NMSE | RMSE | NMSE | RMSE | NMSE |
| Kriging | 0.2537 | 0.7305 | 0.2338 | 0.5412 | 0.2280 | 0.5535 | 0.1993 | 0.4258 |
| RBF | 0.2069 | 0.4054 | 0.1966 | 0.4238 | 0.1898 | 0.4129 | 0.1969 | 0.3920 |
| Deep AE | 0.0828 | 0.0850 | 0.0743 | 0.0694 | 0.0704 | 0.0623 | 0.0645 | 0.0523 |
| AE | 0.0933 | 0.0975 | 0.0723 | 0.0603 | 0.0533 | 0.0328 | 0.0453 | 0.0238 |
| Unet | 0.0249 | 0.0089 | 0.0180 | 0.0043 | 0.0145 | 0.0027 | 0.0129 | 0.0020 |
| RadioUnet | 0.0233 | **0.0082** | 0.0167 | 0.0038 | 0.0134 | 0.0023 | 0.0119 | 0.0017 |
| ACT-GAN | **0.0227** | 0.0086 | **0.0155** | **0.0033** | **0.0124** | **0.0018** | **0.0106** | **0.0013** |

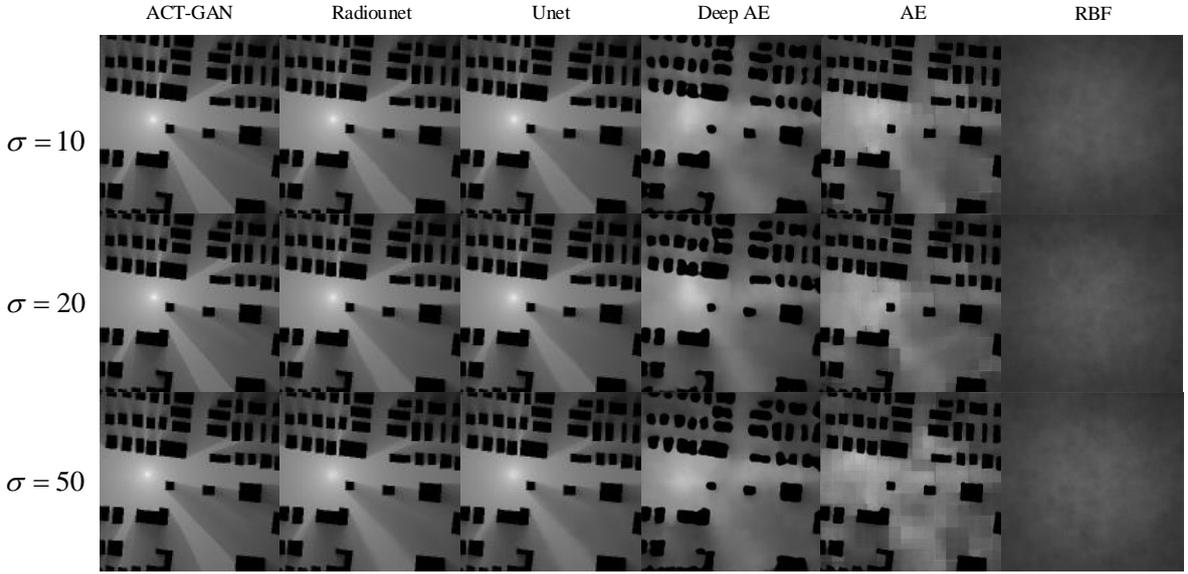

Fig.9. The prediction results under different noise standard deviation

The transmitter location is estimated by finding the maximum pixel point in the constructed radio map. The prediction accuracy presented by mean Euclidean distance is calculated by

$$d = \frac{\sum_{i=1}^{N}\sqrt{\left(\psi_{\arg\max} - \hat{\psi}_{\arg\max}\right)_i^2 \big|_{condition}}}{N} \quad (9)$$

where $\psi_{\arg\max}$ and $\hat{\psi}_{\arg\max}$ are the indexes of the maximum pixel value positions in the label image and the predicted image, respectively, $N$ is the number of samples in the test set Due to the weak ability of some models to fit the data, it will lead to local single pixel values of 255, but it is obvious that these points are not the correct emission source locations. Therefore, in this paper, an index condition is introduced in Eq. (9), which can be expressed as Eq. (10)

$$condition = Max\left(\sum_i^{i+4}\sum_j^{j+4} p_{(i,j)}\right) \quad (10)$$

where $p_{(i,j)}$ represents the receiving field strength (RSS) value for grid $(i, j)$. The prediction accuracy is compared with the different models, as shown in Figure 10.

From Figure 10, it is evident that the localization accuracy based on ACT-GAN is much better than the other models, and when the number of observation points is 150, the average offset of the transmitter position is 0.88 meters.

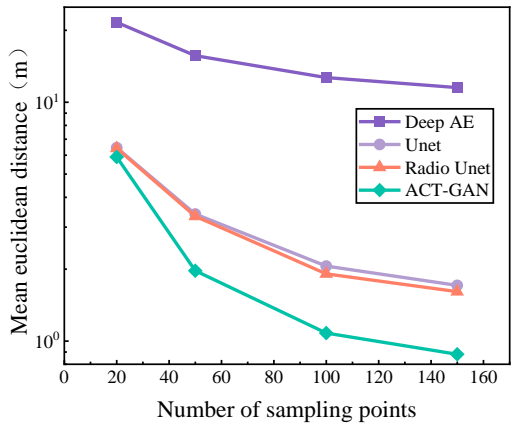

Fig.10. Accuracy of source location prediction.

## 5. Conclusions and future work

In this paper, ACT-GAN which applies the AOT block, CBAM block and T-Conv block to the generator is proposed for radio map construction. Experiments are carried out in three different scenarios, and compared with the other radio map construction models. It is demonstrated that in the scenario without and with sparse discrete observations, the RMSE is reduced by 14.6% and 13.2%, respectively. Interestingly, in the scenario with unknown transmitter locations, the ACT-Gan exhibits high robustness in radio map construction and superior transmitter localization accuracy, providing support for further research into high-quality radio map construction and location applications.

When designing the ACT-GAN structure, we concatenate the AOT module at the full-dimensional level of the encoder, although the larger expansion rate ensures the global accuracy of the constructed map, an excessively large convolutional kernel could create information confusion when scanning images, resulting in the presence of a very small number of white pixel dots in the radio maps with a high density of buildings, which needs to be improved in the future research. In addition, the current study about radio map construction is carried out on a two-dimensional plane, which loses a lot of information such as buildings and antenna heights compared to the actual scene, so that the three-dimensional radio map construction will be our future work.